\documentclass[conference]{IEEEtran}
\IEEEoverridecommandlockouts
\usepackage{afterpage}
\usepackage{float}
\usepackage{placeins}
\usepackage{url}
\usepackage{amsmath,amssymb,amsfonts}
\usepackage{algorithmic}
\usepackage{graphicx}
\usepackage{stfloats} 
\usepackage{textcomp}
\usepackage{xcolor}
\usepackage{rotating}
\usepackage[font=footnotesize]{caption}
\usepackage{subcaption}

\usepackage[numbers,sort&compress,sectionbib]{natbib}
\usepackage{hyperref}

\def\BibTeX{{\rm B\kern-.05em{\sc i\kern-.025em b}\kern-.08em
    T\kern-.1667em\lower.7ex\hbox{E}\kern-.125emX}}

\begin{document}

\title{Utilizing Multilingual Encoders to Improve Large Language Models for Low-Resource Languages%
}

\author{
\IEEEauthorblockN{%
Imalsha Puranegedara\IEEEauthorrefmark{1},
Themira Chathumina\IEEEauthorrefmark{1},
Nisal Ranathunga\IEEEauthorrefmark{1}, \\
Nisansa de Silva\IEEEauthorrefmark{1},
Surangika Ranathunga\IEEEauthorrefmark{2},
Mokanarangan Thayaparan\IEEEauthorrefmark{3}}

\IEEEauthorblockA{\IEEEauthorrefmark{1}Dept.\ of Computer Science \& Engineering, University of Moratuwa, Sri Lanka.\\
\{imalsha.20, themira.20, nisal.20, NisansaDds\}@cse.mrt.ac.lk}

\IEEEauthorblockA{\IEEEauthorrefmark{2}School of Mathematical and Computational Sciences, Massey University, Auckland, New Zealand.\\
s.ranathunga@massey.ac.nz}

\IEEEauthorblockA{\IEEEauthorrefmark{3}Department of Computer Science, The Open University, Milton Keynes, United Kingdom.\\
mokanarangan.thayaparan@open.ac.uk}
}

\maketitle

\begin{abstract}

Large Language Models (LLMs) excel in English, but their performance degrades significantly on low-resource languages (LRLs) due to English-centric training. While there are methods to align LLMs with multilingual encoders such as the Massively Multilingual Text-to-Text Transfer Transformer (mT5), they typically use only the final encoder layer. We propose a novel architecture that fuses all intermediate layers, enriching the linguistic information passed to the LLM. Our approach features two strategies: (1) a Global Softmax weighting for overall layer importance, and (2) a Transformer Softmax model that learns token-specific weights. The fused representations are mapped into the LLM's embedding space, enabling it to process multilingual inputs. The model is trained only on English data, without using any parallel or multilingual data. Evaluated on XNLI, IndicXNLI, Sinhala News Classification, and Amazon Reviews, our Transformer Softmax model significantly outperforms the existing baseline. We observe strong performance gains in LRLs, improving Sinhala classification accuracy from 71.66\% to 75.86\% and achieving clear improvements across Indic languages such as Tamil, Bengali, and Malayalam. These specific gains contribute to an overall boost in average XNLI accuracy from 70.36\% to 71.50\%. This approach offers a scalable, data-efficient path toward more capable and equitable multilingual LLMs\footnote{Our code is available at \url{https://github.com/ImalshaD/MultiLingualImprove}}.
\end{abstract}


\section{Introduction}

Large Language Models (LLMs) are neural networks trained on large corpora to generate and understand human languages. Recent advances favor decoder-only transformer architectures (e.g., GPT-3~\cite{brown2020language}, PaLM~\cite{chowdhery2023palm}, DeepSeek-V3~\cite{puspitasari2025deepseek}) over earlier encoder-based (BERT~\cite{devlin2019bert}) and encoder-decoder models (BART~\cite{lewis2019bart}, T5 \cite{raffel2020exploring}), due to their ability to scale effectively~\cite{kaplan2020scaling}, excel in next-token generation, and generalize well in unseen tasks~\cite{wei2022chain}. Despite their strengths, LLMs remain heavily English-centric, yielding strong performance on reasoning and instruction tasks in English and other high-resource languages~\cite{yu2023metamath,bai2025qwen2}, but performing poorly in low resource languages (LRLs)~\cite{qin2023cross}.

In contrast, multilingual encoders such as XLM-R~\cite{conneau2019unsupervised}, mBERT~\cite{libovicky2019language}, and mT5 encoder~\cite{xue2020mt5} are pretrained on multilingual corpora and are designed for cross-lingual generalization. They generate contextualized token embeddings and are widely used in multilingual Natural Language Processing (NLP) tasks~\cite{choi2021analyzing}. While downstream applications often use only the final encoder layer, recent studies show that intermediate layers also encode valuable syntactic and semantic features~\cite{wang2020multi}, which can enrich cross-lingual transfer when properly utilized. However, these models lack the robust reasoning abilities of English-centric LLMs due to limited exposure to multilingual instruction data.

This mismatch results in a performance gap in multilingual instruction-following~\cite{li2025xifbench}. Attempts to bridge this include: (1) fine-tuning LLMs on translated instruction data~\cite{zhu2024question,ebrahimi2024zero}, (2) using translation at inference time~\cite{shi2022language}, or (3) aligning multilingual encoder outputs (typically final-layer only) with English LLMs~\cite{yoon2024langbridge,huang2024mindmerger}. However, translation-based data creation suffers from noise and semantic drift in LRLs~\cite{chen2023breaking,goyal2022flores}, while multi-stage translation inference introduces latency and compounding errors~\cite{costa2022no,goyal2022flores}. Methods such as LangBridge~\cite{yoon2024langbridge} and MindMerger~\cite{huang2024mindmerger} align multilingual encoders with LLMs using only the final-layer hidden states to understand multilingual input through language agnostic representations provided by multilingual encoders. Such approaches overlook the rich, language-agnostic representations embedded in the intermediate layers of multilingual encoders~\cite{escolano2019multilingual}.

To address these limitations, we propose a method that feeds all multilingual encoder layers into the LLM using a learned, weighted fusion by extending LangBridge architecture. Unlike prior work that uses static final-layer outputs~\cite{yoon2024langbridge}, our model learns to adaptively combine representations across all layers, capturing diverse linguistic features and improving interpretability.
Our methodology targets multilingual understanding and reasoning tasks where the input is multilingual (English and non-English) and the LLM generated output is in English. We train the model exclusively on English instruction data and evaluate it in a zero-shot\footnote{In this context, zero-shot means that the model is trained only on English-language data but is evaluated directly on the same task in languages it has not encountered during training.}  setting on other languages without any multilingual fine-tuning. This setup reflects real-world scenarios where task-specific instruction data is not available in low-resource languages. 

In our experiments, we evaluate the proposed method on a diverse set of datasets covering multiple languages, including XNLI \cite{conneau2018xnli} (15 languages), IndicXNLI \cite{aggarwal2022indicxnli} (12 Indic languages), Sinhala News Classification \cite{de2019survey,de2015sinhala}, and Amazon Multilingual Reviews \cite{hou2024bridging} (covering languages such as English, Spanish, French, German, Chinese, and Japanese). 
Empirically, we observe that our approach yields consistent improvements across languages, with notably higher gains for languages whose scripts are similar to English. The ability to leverage shared script and structural similarities enables more effective transfer, offering a practical benefit for expanding LLM capabilities to low-resource, script-similar languages without requiring additional annotated data.

\section{Related Work}

\subsection{Aligning Pretrained Representations}

Aligning pretrained encoder outputs with LLMs has gained traction in multi-modal tasks (e.g., vision-language models~\cite{liu2023visual}) and is now being extended to multilingual settings. Approaches such as 
LangBridge~\cite{yoon2024langbridge} and MindMerger~\cite{huang2024mindmerger} aim to equip English-centric LLMs with multilingual capabilities by incorporating outputs from multilingual encoders.
LangBridge uses a soft replacement strategy, projecting the multilingual encoder’s final-layer hidden states into the LLM’s input space via a learned transformation. MindMerger introduces a two-stage process: (1) a mapping stage using bilingual pairs to align multilingual representations with the LLM, and (2) an augmentation stage to integrate them with the LLM’s inputs.
However, both methods rely solely on the final-layer encoder output, neglecting rich linguistic information present in intermediate layers~\cite{escolano2019multilingual}. Moreover, MindMerger depends on parallel instruction or bilingual data, which is scarce for many low-resource languages, limiting its applicability in zero-shot scenarios.
%

\subsection{Using All Layers of the Encoder}

Prior work shows that intermediate encoder layers capture diverse linguistic features useful for downstream tasks. Methods such as weighted summation or attention-based fusion~\cite{wang2020multi} have been applied in English classification and Neural Machine Translation. For example, Bapna et al.\cite{bapna2018training} use a trainable softmax over encoder layers for neural machine translation, while Liu et al.\cite{liu2020understanding} introduce fixed, learned scaling vectors per layer. DWAtt~\cite{elnokrashy2022depth} further improves layer selection through an attention mechanism.
These strategies have proven effective in monolingual or bilingual contexts, but have not been extended to multilingual instruction-following LLMs, especially where input is non-English and output is English.

\section{Methodology}

We propose an architecture that enhances multilingual understanding in English-centric LLMs by extending the LangBridge~\cite{yoon2024langbridge} architecture. Unlike LangBridge, which uses only the final-layer hidden state from a multilingual encoder, our method integrates all intermediate layer representations through a learned fusion mechanism. This allows the model to dynamically capture both low-level and high-level linguistic features, improving its ability to understand and reason over inputs in diverse languages, particularly in zero-shot, low-resource scenarios.

Given a multilingual instruction prompt, we first pass it through a frozen multilingual encoder, which outputs token-level hidden states from all transformer layers. These multi-layered representations encode different linguistic abstractions, from syntactic signals in early layers to semantic and task-relevant features in deeper ones. To effectively leverage this full depth, we introduce a trainable depth-wise weighting mechanism that learns to assign importance scores to each layer per token, generating a single fused representation per token.
These fused token embeddings are then projected into the input space of an English-centric LLM using a learned linear transformation. The encoder and LLM remain frozen during training; however, the fusion and projection components are trained end-to-end using only English instruction–response data. This setup enables the model to align multilingual inputs with English LLM reasoning capabilities without the need for multilingual fine-tuning or parallel instruction data.




\begin{figure*}[!t]
  \centering
  \subfloat[\hspace{1.5cm} (a) LangBridge\label{fig:arch-lang}]{%
    \includegraphics[
        width = 0.25\textwidth,
        height= 0.32\textheight,
        keepaspectratio]{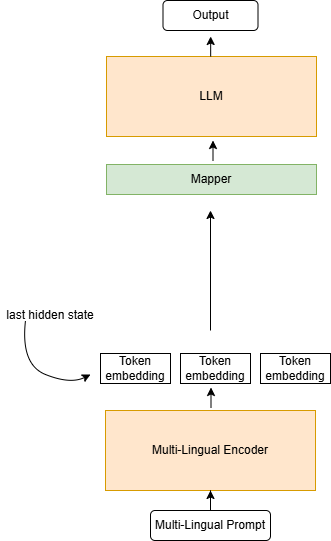}}
  \hfill
  \subfloat[\hspace{-1cm}(b) Proposed\label{fig:arch-prop}]{%
    \includegraphics[
        width = 0.25\textwidth,
        height= 0.32\textheight,
        keepaspectratio]{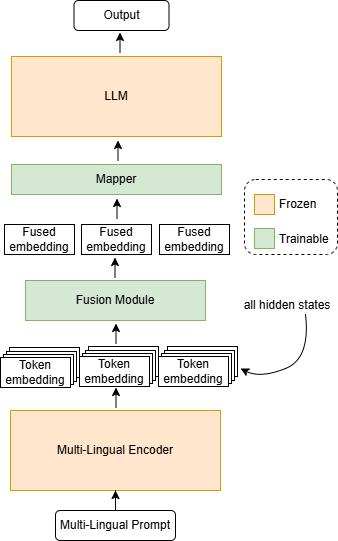}}
  \hfill
  \subfloat[(c) Training\label{fig:arch-train}]{%
    \includegraphics[
        width = 0.25\textwidth,
        height= 0.32\textheight,
        keepaspectratio]{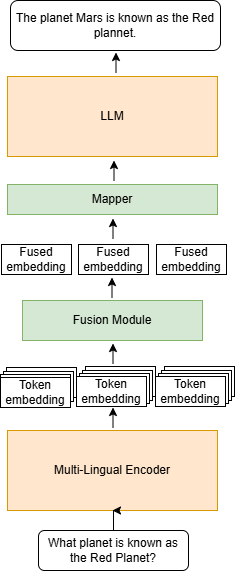}}
  \hfill
  \subfloat[(d) Inferencing\label{fig:arch-inf}]{%
    \includegraphics[
        width = 0.25\textwidth,
        height= 0.32\textheight,
        keepaspectratio]{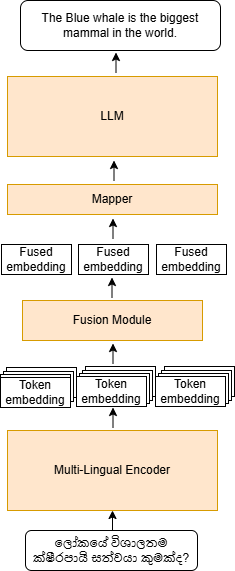}}
  
  \caption{Figure 1: Model comparison and usage.
            (a) LangBridge uses only the final encoder layer.
            (b) Our approach fuses all hidden layers via a fusion module.
            (c) Training with English inputs.
            (d) Zero-shot inference on a low-resource language.}
  \label{fig:arch}
\end{figure*}

\subsection{Fusion Module}

The Fusion Module combines all intermediate hidden states from the multilingual encoder into a single embedding for each token.
For each token, the multilingual encoder produces a sequence of hidden states 
\( h_1, h_2, \dots, h_L \in \mathbb{R}^d \), 
where \( d \) is the embedding dimension. The Fusion Module computes a weighted sum of these representations:

\begin{equation}
h_{\text{fused}} = \sum_{i=1}^{L} \alpha_i \cdot h_i
\label{eq:fused}
\end{equation}

where \( \alpha_i \) denotes the normalized weight assigned to the \( i \)-th layer, and the sum is performed independently for each token.

We propose two strategies for computing these weights:

\subsubsection{Global Softmax Weighting}

Here, each \( \alpha_l \) is a scalar weight that determines the contribution of the \( l \)-th layer to the final embedding. These weights are shared across all tokens and samples in the batch, and are normalized using a temperature-scaled softmax:

\begin{equation}
\alpha_l = \frac{\exp(\tau \cdot w_l)}{\sum_{k=1}^{L} \exp(\tau \cdot w_k)}
\label{eq:alpha_norm}
\end{equation}

where \( w \in \mathbb{R}^L \) is a trainable parameter vector initialized with small values. The effective temperature \( \tau \) is not directly learned but computed as:

\begin{equation}
\tau = \texttt{base\_temp} + \texttt{temp} \times \texttt{factor}
\end{equation}

where \texttt{base\_temp} is a fixed constant (e.g., \(10^2\)), \texttt{factor} is a large scaling constant (e.g., \(10^5\)), and \texttt{temp} is a learnable parameter. 



\subsubsection{Token-Wise Softmax Weighting}
Rather than applying a uniform weighting across all tokens, we compute a unique layer attention distribution for each token individually. 
The central hypothesis underlying this approach is that different tokens may benefit from information residing in different layers of the multilingual encoder. By leveraging the Encoder-based attention mechanism in the Transformer architecture~\cite{vaswani2017attention}  over the set of a token's layer-wise representations, the model can learn to attend to the most informative layers per token.

This fine-grained, token-specific layer fusion enables the model to dynamically extract semantic features best suited for each input position. Importantly, such a mechanism can be particularly beneficial for languages that share similar scripts with English. Since attention weights are computed at the token level, the model can more effectively align similar subword patterns across languages, thereby enhancing transfer and generalization in multilingual instruction following.


Let \( \mathcal{E} \) be a multilingual encoder with \( L \) layers. Given an input sequence \( X = \{x_1, x_2, \dots, x_T\} \) in a non-English language, the encoder produces for each token \( x_t \) a set of hidden states from all layers:
 
\begin{equation}
\mathbf{H}_t = \{\mathbf{h}_t^{(1)}, \mathbf{h}_t^{(2)}, \dots, \mathbf{h}_t^{(L)}\},
\quad \mathbf{h}_t^{(l)} \in \mathbb{R}^d
\label{eq:hidden_states}
\end{equation}
 
where \( d \) is the hidden dimension of the encoder.
 
 
To learn token-specific importance over layers, we augment \( \mathbf{H}_t \) with a learnable query vector \( \mathbf{c} \in \mathbb{R}^d \), and form the following sequence:
 
\begin{equation}
\mathbf{S}_t = [\mathbf{c}; \mathbf{h}_t^{(1)}; \dots; \mathbf{h}_t^{(L)}] 
              \in \mathbb{R}^{(L+1) \times d}
\label{eq:state_concat}
\end{equation}
 
We then add positional embeddings \( \mathbf{P} \in \mathbb{R}^{(L+1) \times d} \) to \( \mathbf{S}_t \) and pass the result through a Transformer encoder block \( \text{TransEnc}(\cdot) \):
 
\begin{equation}
\mathbf{U}_t = \text{TransEnc}\bigl(\mathbf{S}_t + \mathbf{P}\bigr)
\label{eq:transenc}
\end{equation}

Let \( \mathbf{u}_t^{(0)} \in \mathbb{R}^d \) be the transformed output corresponding to the query vector \( \mathbf{c} \). We then project this vector to a layer attention score vector \( \boldsymbol{\alpha}_t \in \mathbb{R}^L \):
 
\begin{equation}
\boldsymbol{\alpha}_t = \text{softmax}\!\bigl(\mathbf{W}_a \,\mathbf{u}_t^{(0)} + \mathbf{b}_a\bigr)
\label{eq:alpha_t}
\end{equation}
 
where \( \mathbf{W}_a \in \mathbb{R}^{L \times d} \) and \( \mathbf{b}_a \in \mathbb{R}^L \) are trainable parameters. The final token representation \( \tilde{\mathbf{h}}_t \in \mathbb{R}^d \) is computed as the weighted sum over all encoder layers:
 
\begin{equation}
\tilde{\mathbf{h}}_t = \sum_{l=1}^{L} \alpha_t^{(l)} \cdot \mathbf{h}_t^{(l)}
\label{eq:fused_hidden}
\end{equation}

\subsection{Mapping to LLM Embedding Space}
 
Since the multilingual encoder and LLM may operate in different vector spaces, we apply a linear projection \( \mathbf{W}_p \in \mathbb{R}^{d' \times d} \) to map the fused representations into the LLM’s embedding space \( \mathbb{R}^{d'} \):
 
\begin{equation}
\mathbf{z}_t = \mathbf{W}_p \tilde{\mathbf{h}}_t + \mathbf{b}_p
\label{eq:projection}
\end{equation}

The transformed sequence \( \mathbf{Z} = \{\mathbf{z}_1, \dots, \mathbf{z}_T\} \) is then used as input to the frozen LLM decoder.
 
\subsection{Training Objective}
 
We train the fusion module and the projection layer while keeping the encoder and LLM weights frozen. The training is conducted using a prefix language modeling objective: given input \( X \) in a non-English language, the model is trained to generate the corresponding output \( Y = \{y_1, y_2, \dots, y_K\} \) in English. The objective is the cross-entropy loss over the predicted token distribution:
 
\begin{equation}
\mathcal{L}_{\text{CE}} = - \sum_{k=1}^{K} \log p\!\bigl(y_k \mid y_{<k}, \mathbf{Z}\bigr)
\label{eq:ce_loss}
\end{equation}

This setup enables multilingual instruction following a \textit{zero-shot} setting by leveraging pretrained reasoning capabilities of the LLM and language representations of the multilingual encoder.

\section{Experimental Setup}

\subsection{Datasets}

To evaluate the effectiveness of our proposed method, we conduct experiments on a diverse set of multilingual benchmarks that assess natural language understanding, reasoning, and generalization to low-resource languages. We use the XNLI dataset~\cite{conneau2018xnli}. 
To further evaluate performance on low-resource languages, we utilize the IndicXNLI dataset~\cite{aggarwal2022indicxnli}, an extension of XNLI designed for Indian languages with limited resources. Additionally, we include the Sinhala News Classification dataset~\cite{de2015sinhala}, which involves categorizing news headlines written in Sinhala. We also employ the multilingual Amazon Reviews sentiment analysis dataset~\cite{hou2024bridging} to test sentiment classification across multiple languages.

\subsection{Models and Baseline}

For our experiments, we use Llama
3.2 1B Instruct~\cite{meta2023introducing} as the base large language model (LLM). This model has strong instruction-following capabilities in English but lacks robust multilingual understanding due to its English-centric pretraining. As the multilingual encoder, we employ mT5-Large, which is pretrained on a wide range of multilingual data and serves as a strong source of cross-lingual representations. Given the high computational cost associated with large-scale LLMs, we restrict our experiments to smaller-scale models to ensure feasibility and reproducibility.

Our primary baseline is \textbf{LangBridge}~\cite{yoon2024langbridge},
a recent and competitive architecture designed to enhance the multilingual capabilities of LLMs by aligning a multilingual encoder with an English-centric decoder. This was enhanced by Mindmerger~\cite{huang2024mindmerger} with the use of parallel instruction data. In our experiments, we do not consider using parallel instruction data (non-English instruction and English response) because it does not represent real world scenarios where parallel instruction data are limited in non-English languages. Thus Mindmerger is not considered a baseline.
We compare our proposed method against LangBridge using two  variants of our model: Global Softmax Weighting (\textbf{Global Softmax}) and Token-Wise Softmax Weighting (\textbf{Transformer SM}). 
\subsection{Training details}
For each dataset, we train the model for 3 epochs using a learning rate of 3e-5 and a cosine learning rate scheduler. We use only the English subset for training and evaluate the model on both English and other languages. For Global Softmax Weighting, we set the temperature to 1e-2. For Token-Wise Softmax Weighting, we use a single transformer encoder layer with 4 attention heads. All models are trained on a single RTX 4090 GPU with 24 GB of VRAM.

\section{Results and Analysis}

\begin{table*}[!t]
\caption{XNLI Results (accuracy)}
\begin{center}
\begin{tabular}{|l|c|c|c|c|c|c|c|c|c|c|c|c|c|c|c|c|}
\hline
\textbf{Experiment} & \textbf{avg} & \textbf{en} & \textbf{ar} & \textbf{bg} & \textbf{de} & \textbf{el} & \textbf{es} & \textbf{fr} & \textbf{hi} & \textbf{ru} & \textbf{sw} & \textbf{th} & \textbf{tr} & \textbf{ur} & \textbf{vi} & \textbf{zh} \\
\hline
LangBridge & 70.36 & 77.31 & 69.62 & 73.35 & 71.54 & 71.76 & 74.01 & 73.43 & 67.35 & 71.16 & 66.09 & \textbf{67.49} & 68.54 & \textbf{65.37} & 69.80 & 68.58 \\
Global Softmax  & 70.63 & 79.30 & 69.24 & 74.75 & 71.74 & 72.75 & \textbf{75.51} & 74.49 & \textbf{68.14} & 71.32 & 62.81 & 66.69 & 68.20 & 64.45 & \textbf{70.86} & 69.24 \\
Transformer SM & \textbf{71.50} & \textbf{79.42} & \textbf{70.34} & \textbf{74.99} & \textbf{73.87} & \textbf{73.47} & 75.43 & \textbf{75.55} & 68.02 & \textbf{72.22} & \textbf{67.27} & 67.01 & \textbf{69.56} & 64.95 & 70.60 & \textbf{69.80} \\
\hline
\end{tabular}
\label{tab:xnli_results}
\end{center}
\end{table*}

\begin{table*}[!t]
\caption{INDIC XNLI Results (accuracy)}
\begin{center}
\begin{tabular}{|l|c|c|c|c|c|c|c|c|c|c|c|c|}
\hline
\textbf{Experiment} & \textbf{avg} & \textbf{as} & \textbf{bn} & \textbf{gu} & \textbf{hi} & \textbf{kn} & \textbf{ml} & \textbf{mr} & \textbf{or} & \textbf{pa} & \textbf{ta} & \textbf{te} \\
\hline
LangBridge          & 65.75 & 62.69 & 68.70 & 67.21 & 70.54 & 67.29 & 66.17 & 64.07 & 57.05 & 66.41 & 66.63 & 66.49 \\
Global Softmax      & \textbf{66.96} & \textbf{65.03} & 69.08 & \textbf{68.94} & \textbf{71.38} & 68.44 & 66.59 & \textbf{65.27} & \textbf{58.40} & \textbf{69.34} & 66.73 & 67.37 \\
Transformer SM      & 66.94 & 64.17 & \textbf{69.70} & 68.66 & 71.26 & \textbf{69.48} & \textbf{68.06} & 64.95 & 55.11 & 68.96 & \textbf{67.50} & \textbf{68.52} \\
\hline
\end{tabular}
\label{tab:indic_results}
\end{center}
\end{table*}

\begin{table}[!htb]
\caption{News Results (accuracy)}
\begin{center}
\begin{tabular}{|l|c|c|}
\hline
\textbf{Experiment} & \textbf{en} & \textbf{si} \\
\hline
LangBridge     & 88.23 & 71.66 \\
Global Softmax & \textbf{88.97} & 73.24 \\
Transformer SM & 88.79 & \textbf{75.86} \\
\hline
\end{tabular}
\label{tab:news_results_final}
\end{center}
\end{table}


\begin{table}[!htb]
\caption{Amazon Review Sentiment Analysis (accuracy)}
\centering
\footnotesize
\setlength{\tabcolsep}{3pt} 
\resizebox{\columnwidth}{!}{%
\begin{tabular}{|l|c|c|c|c|c|c|c|}
\hline
\textbf{Experiment} & \textbf{en} & \textbf{es} & \textbf{fr} & \textbf{de} & \textbf{zh} & \textbf{ja} & \textbf{avg} \\
\hline
LangBridge     & 64.62 & 54.84 & 55.40 & \textbf{61.04} & 50.64 & 51.60 & 56.36 \\

Global Softmax        & 64.92 & 55.22 & \textbf{56.60} & 60.9 & \textbf{51.26} & 51.66 & 56.76 \\

Transformer SM & \textbf{65.06} & \textbf{55.72} & 56.00& 60.80 & 50.90 & \textbf{52.46} & \textbf{56.82} \\
\hline
\end{tabular}}
\label{tab:amazon_sentiment_results}
\end{table}

\begin{table}[!htb]
\caption{Inference Latency and GPU Memory Usage.}
\begin{center}
\begin{tabular}{|l|c|c|}
\hline
\textbf{Model/Method} & \textbf{Latency ms per token} & \textbf{Peak VRAM MB} \\
\hline
Translation pipeline & 234.68 & \textbf{5000.03} \\
Langbridge & 24.23 & 11453.35 \\
Global Softmax & \textbf{23.74} & 11453.51 \\
Transformer SM & 25.41 & 11535.03 \\
\hline
\end{tabular}
\label{tab:gpu_latency_efficiency}
\end{center}
\end{table}

Our evaluation across multiple multilingual benchmarks reveals that integrating intermediate encoder layers through learned fusion significantly enhances the performance of English-centric LLMs, especially in zero-shot scenarios involving low-resource languages.

\subsection{Overall Performance}

Both of our proposed models, Global Softmax and Transformer SM, consistently outperform the LangBridge baseline in all evaluated data sets. As shown in Table~\ref{tab:xnli_results}, Transformer SM achieves the highest average accuracy on the XNLI benchmark, with improvements evident in both high-resource (e.g., English, French) and non-English languages. This trend persists across IndicXNLI (Table~\ref{tab:indic_results}), Sinhala News Classification (Table~\ref{tab:news_results_final}), and Amazon Review Sentiment Analysis (Table~\ref{tab:amazon_sentiment_results}), demonstrating strong generalization and transfer capabilities.

\subsection{Improvements in Low-Resource Languages}

The most significant gains are observed in LRLs, which are traditionally underserved by English-centric LLMs. In the IndicXNLI benchmark (Table~\ref{tab:indic_results}), both fusion strategies outperform LangBridge in nearly all languages, with marked improvements in Assamese, Gujarati, and Malayalam. Similarly, on the Sinhala News Classification task (Table~\ref{tab:news_results_final}), Transformer SM notably boosts performance over LangBridge, indicating its ability to capture meaningful linguistic representations even in limited-resource settings.
These results validate our core hypothesis that incorporating representations from all encoder layers provides richer linguistic grounding, particularly crucial for LRLs where single-layer representations may not be sufficient to convey complex linguistic structure.

\subsection{Token-Wise Fusion vs. Global Weighting}


While Global Softmax improves over LangBridge in most settings, Transformer SM often yields the best results, particularly in morphologically rich and structurally diverse languages such as Tamil, Malayalam, Arabic, and Russian. This suggests that the token-wise fusion mechanism offers finer control over the alignment process, enabling the model to dynamically select the most relevant layers on a per-token basis. The advantage is especially visible in the XNLI (Table~\ref{tab:xnli_results}) and Indic XNLI (Table~\ref{tab:indic_results}) datasets, where Transformer SM either matches or exceeds the performance of Global Softmax across languages.

\subsection{Cross-Task Generalization}

Our architecture generalizes well beyond natural language inference. In sentiment analysis (Table~\ref{tab:amazon_sentiment_results}), Transformer SM again demonstrates consistent improvements over LangBridge, particularly in non-English languages such as Spanish, French, and Japanese. This illustrates the effectiveness of our approach across diverse NLP task types, confirming that the enriched encoder fusion improves not just language understanding but also task-specific reasoning.

\subsection{Layer Weight Distribution Analysis}

\begin{figure}[!htbp]
\centering
\includegraphics[width=\columnwidth]{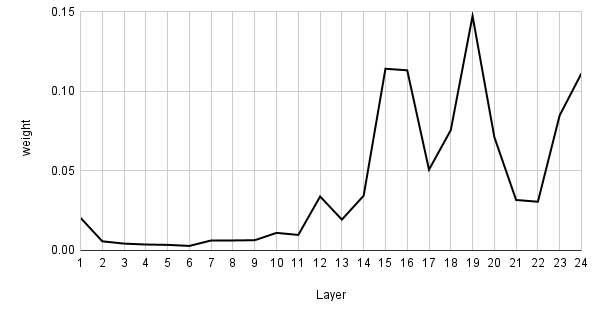} 
\caption{Comparison of layer-wise contributions and weight distribution in the global softmax mechanism during the XNLI experiment.}
\label{fig-softmax}
\end{figure}

To better understand the model’s behavior, we visualize the learned layer weights in Figure~\ref{fig-softmax}. The plot reveals that lower layers contribute minimally, while middle and upper layers receive significantly higher weights, peaking around layers 15–19. This pattern is consistent with prior findings
that intermediate encoder layers often encode critical syntactic and semantic features \cite{choenni2023examining}. 

\subsection{Inference Efficiency Analysis}
Our efficiency analysis on the Sinhala News Classification dataset (Table~\ref{tab:gpu_latency_efficiency}) reveals a key trade-off. A baseline \textit{Translation pipeline} using NLLB 1.3B translation model has low memory usage but suffers from high latency. Conversely, the fusion-based methods (`LangBridge`, `Global Softmax`, and `Transformer SM`) show comparably low latencies, making them significantly more time-efficient for real-time applications despite their higher VRAM requirements.


\section{Conclusion}


This work introduces a fusion-based method to enhance the multilingual capabilities of English-centric Large Language Models (LLMs) by incorporating all hidden layers of a frozen multilingual encoder. Unlike prior approaches that rely only on the final layer, our method uses either global or token-wise fusion to generate richer token embeddings, trained using a prefix language modeling objective without requiring parallel multilingual data.
Experiments on different datasets show consistent improvements, especially in LRLs. The Transformer Softmax variant achieved the highest gains. These results confirm the value of deep encoder fusion in zero-shot multilingual settings. Thus, our findings highlight the value of multilayer encoder fusion in bridging performance gaps for LRLs and offer a scalable path to more inclusive multilingual NLP.
However, due to computational constraints, the evaluations were limited to an LLM with 1B parameters. It remains to be tested whether our method provides similar gains when applied to larger or already multilingual-tuned LLMs such as Mistral-MoE. This is a key direction for future research.

\bibliographystyle{IEEEtranN}
\bibliography{references}

\end{document}